\documentclass{article}

\usepackage{PRIMEarxiv}

\usepackage{multirow}       
\usepackage[utf8]{inputenc} 
\usepackage[T1]{fontenc}    
\usepackage{url}            
\usepackage{booktabs}       
\usepackage{amsfonts}       
\usepackage{nicefrac}       
\usepackage{microtype}      
\usepackage{fancyhdr}       
\usepackage{graphicx}       
\usepackage{natbib}
\usepackage{array}
\usepackage{amsmath}
\usepackage{mathabx}
\usepackage{bigfoot}
\usepackage{amsmath}
\usepackage{amsfonts}
\usepackage{algorithm}
\usepackage{algpseudocode}
\usepackage{pifont}
\usepackage{cleveref}
\usepackage{caption}
\usepackage{graphicx}
\usepackage{footnote}

\title{Science Out of Its Ivory Tower: Improving Accessibility with Reinforcement Learning
}

\author{
  Haining Wang\\
  Indiana University\\
  Bloomington, Indiana, USA\\
  \texttt{hw56@indiana.edu}\\
  \And
  Jason Clark\\
  Montana State University\\
  Bozeman, Montana, USA\\
  \texttt{jaclark@montana.edu}\\
  \And
  Hannah McKelvey\\
  Montana State University\\
  Bozeman, Montana, USA\\
  \texttt{hannah.mckelvey@montana.edu}\\
  \And
  Leila Sterman\\
  Montana State University\\
  Bozeman, Montana, USA\\
  \texttt{leila.sterman@montana.edu}\\
  \And
  Zheng Gao\\
  Ant Group\\
  Sunnyvale, California, USA\\
  \texttt{zheng.gao@antgroup.com}\\
  \And
  Zuoyu Tian\\
  Macalester College\\
  Saint Paul, Minnesota, USA\\
  \texttt{ztian@macalester.edu}\\
  \And
  Sandra Kübler\\
  Indiana University\\
  Bloomington, Indiana, USA\\
  \texttt{skuebler@iu.edu}\\
  \And
  Xiaozhong Liu\\
  Worcester Polytechnic Institute\\
  Worcester, Massachusetts, USA\\
  \texttt{xliu14@wpi.edu}\\
}

\begin{document}
\maketitle

\begin{abstract}
A vast amount of scholarly work is published daily, yet much of it remains inaccessible to the general public due to dense jargon and complex language. To address this challenge in science communication, we introduce a reinforcement learning framework that fine-tunes a language model to rewrite scholarly abstracts into more comprehensible versions. Guided by a carefully balanced combination of word- and sentence-level accessibility rewards, our language model effectively substitutes technical terms with more accessible alternatives, a task which models supervised fine-tuned or guided by conventional readability measures struggle to accomplish. Our best model adjusts the readability level of scholarly abstracts by approximately six U.S. grade levels---in other words, from a postgraduate to a high school level. This translates to roughly a 90\% relative boost over the supervised fine-tuning baseline, all while maintaining factual accuracy and high-quality language. An in-depth analysis of our approach shows that balanced rewards lead to systematic modifications in the base model, likely contributing to smoother optimization and superior performance. We envision this work as a step toward bridging the gap between scholarly research and the general public, particularly younger readers and those without a college degree.
\end{abstract}


\section{Introduction}

\begin{figure*}[!ht]
  \centering
  \includegraphics[scale=0.6]{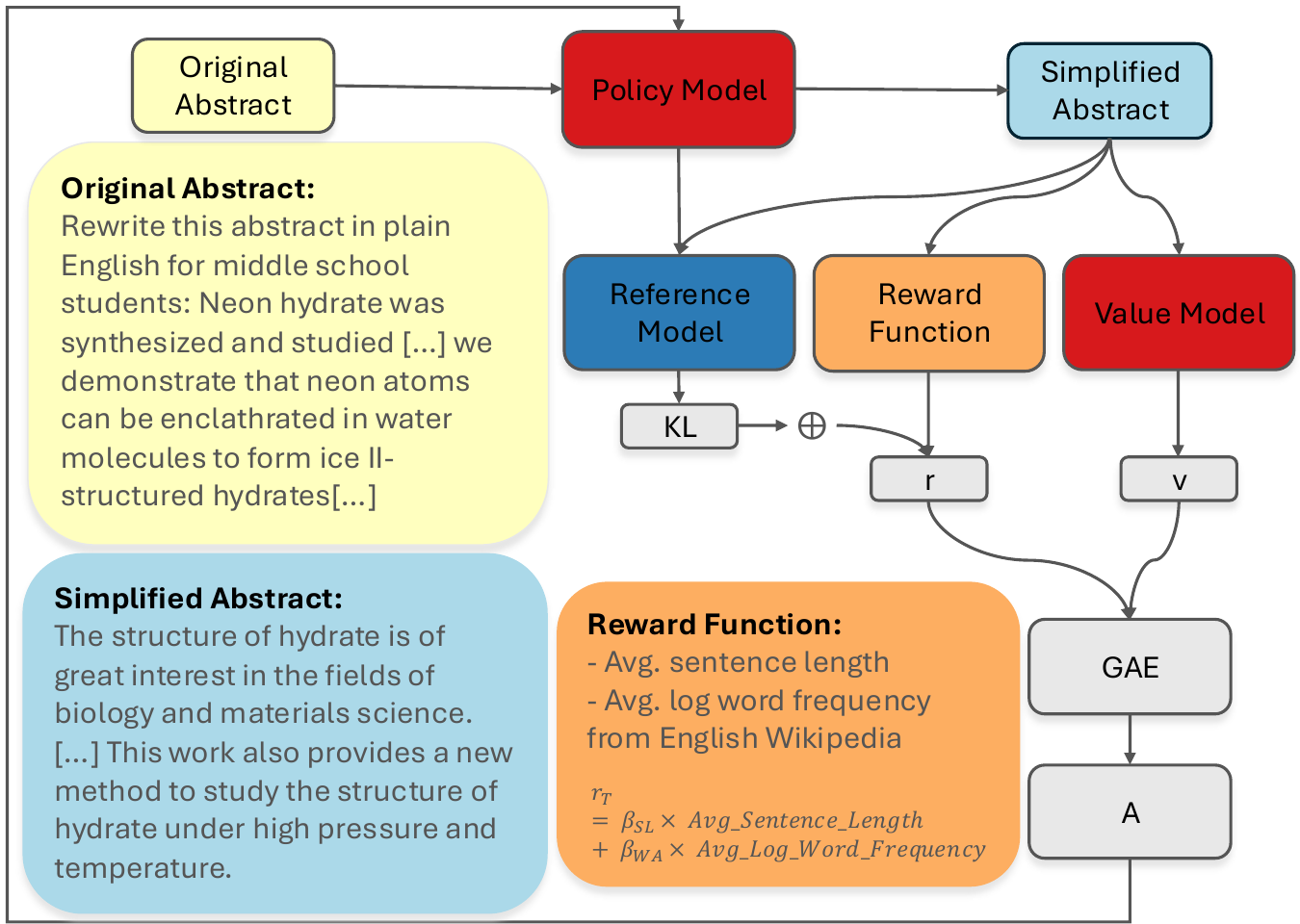}
\caption{RLAM Training Workflow: RLAM rewrites scholarly abstracts using PPO (Section~\ref{sec: RLAM}), guided by a reward function that balances average sentence length and word accessibility. The policy model is optimized iteratively through an actor-critic framework: it generates simplified abstracts, whose quality is assessed by the reward function and regularized through KL divergence from the frozen reference model (the supervised fine-tuning model). The reward signal is further contrasted with expected returns estimated by the value model, implemented as a linear head atop the policy model. The resulting advantage is distributed across tokens via Generalized Advantage Estimation (GAE). RLAM enables the lightweight Gemma-2B model to reduce abstracts from postgraduate to high school readability, achieving a 90\% improvement over the supervised baseline.}
\label{fig: flowchart}
\end{figure*}

\subsection{Accessible Language in Science Communication}
At first glance, the daily publication of tens of thousands of scientific papers---many freely accessible through open science and open access initiatives---suggests few barriers to knowledge dissemination. 
However, two key facts challenge this perception and show that significant barriers remain. 
A recent survey by the U.S. Department of Education found that more than half of U.S. adults aged 16 to 74 (54\%, or 130 million people) read at or below a sixth-grade level \citep{rothwell2020assessing}. 
Meanwhile, an analysis of the readability of biomedical research abstracts published from 1881 to 2015 found that scientific writing has become increasingly hard to read over time \citep{plavensigray2017readability}. 
Even when intended to be accessible, scientific abstracts typically require a postgraduate level of reading comprehension due to jargon use and sentence structure \citep{wang2025simplifying}.
This discrepancy leaves a significant portion of the population---including young readers and adults without advanced degrees---unable to fully engage with scientific works, even if these are made freely available online.
The ``infodemic'' surrounding COVID-19 highlighted this issue: the urgent need for understandable information about the virus clashed with the complex presentation of scientific findings---leading many to turn to more digestible but less reliable narratives on social media \citep{wang2019systematic, islam2020misinformation, calleja2021public}. 

While the legal and medical fields have long been encouraged to use accessible language as a clear conduit for public engagement \citep{mazur2000revisiting, petelin2010considering}, momentum for the adoption of accessible language within scientific communities has been building, roughly since the start of the open science movement \citep{schriver2017plain}.
For instance, the National Institutes of Health (NIH) advocates for ``clear and simple'' principles when communicating with audiences with limited health literacy, and the \emph{Proceedings of the National Academy of Sciences of the United States of America} (\emph{PNAS}) requires authors to submit a significance statement accessible to non-experts \citep{berenbaum2021covid, pool2021infodemic}.
However, there are inherent conflicts between the specialized nature of communication among disciplinary peer scholars and the public-oriented dissemination of scientific findings. 
Even assuming that communicating scholarly works in plain language is possible, it will inevitably increase the communication cost among domain experts and create confusion at the more advanced levels, compared to the use of jargon and technical terms.
In an era of increasingly specialized scientific research, this conundrum is not easily addressed by scientists or disseminators.
Moreover, expecting individual researchers to translate their findings for the public imposes an unrealistic burden, given their primary focus on advancing specialized knowledge.
At the same time, public demand for clarity remains high, especially during fast-moving scientific crises where accessible information is urgently needed.

In response, we propose addressing the need for communicating scientific findings to a broader audience by \emph{rewriting scholarly abstracts with simpler words and grammar using language models.} 
Since readability is key to comprehending scholarship \citep{flesch1946art, duBay2004principles, kerwer2021straight}, we envision the resulting accessible narratives as paving the way for the ``last mile'' of science, broadening access to scientific understanding and engagement, especially for younger readers and those without a college degree.

\subsection{Challenges to Effective Simplification}\label{sec: simplification_challenges}
Fine-tuning a language model using pairs of abstracts and their accessible versions is the \emph{de facto} method for automating the rewriting of scholarly abstracts into more accessible versions \citep{xu2015problems, goldsack2022making, joseph2023multilingual}. 
Accordingly, we introduced the Scientific Abstract-Significance Statement (SASS) corpus \citep{wang2025simplifying}, a dataset composed of paired abstracts and significance statements from diverse disciplines, with the latter targeting ``an undergraduate-educated scientist outside their field of specialty'' \citep{berenbaum2021covid, pool2021infodemic}.
Although the simplified abstracts generated from language models fine-tuned on the SASS corpus are approximately three grade levels more readable than the original abstracts, as measured by U.S. grade-based readability scores \citep[Sec.~6]{wang2025simplifying}, the documents are still not sufficiently accessible; even the best models produce college-level texts. 
Additionally, because the vocabulary used in significance statements is often just as complex as that found in the abstracts themselves \citep[Sec.~3]{wang2025simplifying}, the readability improvements are primarily due to shorter sentences, and technical terms remain inadequately addressed.

Alternatively, the optimization of a language model can be guided by a chosen objective in an actor-critic manner \citep{ramamurthy2023is}. 
It is intuitive to choose an established document readability measure, such as the Automated Readability Index (ARI; see Section~\ref{eval: accessibility}), to assess the overall readability of the outputs generated by the language model.\footnote{The objective can also be reading difficulties ranked by human raters, similar to reinforcement learning from human feedback (RLHF) \citep{ouyang2022training}, where the feedback focuses specifically on document readability rather than general preferences. While this type of feedback is of high quality, it is significantly more expensive than our reward functions. 
Additionally, document readability is more tangible and therefore easier to quantify than the more ambiguous concept of ``human preference.''
For these reasons, we did not consider using human raters in this study.} 
However, we found that the optimization of language models guided by ARI is highly unstable, often resulting in the production of seemingly more accessible versions that still contain many technical terms.
Inspired by \citet{riddell2021varieties}, we decomposed the measurement of document readability into two distinct measures: one at the sentence level and one at the word level. 
We then prioritized word-level accessibility in the optimization to encourage the model to use more accessible words instead of simply shortening sentences.

\subsection{Contribution}
Our work aims to serve as a bridge between scholarly works and the general public, particularly benefiting younger readers and those without a college degree.

\begin{enumerate}
    \item We address the common challenges in science communication by rewriting scholarly abstracts at a high school reading level using a language model.
    \item We identify the challenges language models face in properly addressing jargon and propose Reinforcement Learning from Accessibility Measures (RLAM) as a means to improve the models' use of accessible terms in their rewrites. RLAM-trained language models can significantly reduce the reading level of a scholarly abstract from a postgraduate level to a high school level, achieving a 3-grade-level reduction---or about a 90\% performance boost---compared to models fine-tuned using the same corpus.
    \item We observe systematic differences between reinforcement learning models guided by different rewards and conclude that disproportionate weights for sentence-level rewards contribute to unstable training and lower simplification quality.
\end{enumerate}

Our code, training logs, and model generations are available at \url{https://github.com/Wang-Haining/RLAM} under a permissive license.

\section{Literature Review}

\subsection{The Common Challenges of Science Communication}
Science communication research uses several models to describe strategies for public understanding of and engagement with science, including the deficit model, the dialogue model, and the participatory model \citep{trench2008towards, hetland2014models, sokolovska2019communication}.
The deficit model assumes that the public lacks scientific knowledge and that science communication should focus on simplifying complex scholarship through one-way information transmission. 
The dialogue model encourages two-way communication, where scientists and the public engage in conversations that foster mutual understanding \citep{trench2008towards, joly2008lost}. 
The participatory model involves the public as active participants in the scientific process, recognizing their contributions as equally valuable to those of scientists \citep{brossard2009critical}. 
In practice, the dialogue model suggests that researchers should engage with the public through mediums such as social media \citep{davies2008constructing, hara2019emerging, knox2021public}, by crafting more digestible manuscripts in research \citep{maurer2021lessons}, and by developing plain language guidelines for practitioners \citep{grene2017use}. 
Communicators are encouraged to scaffold content by providing the audience with relevant, easily understandable material that either gradually increases in complexity or includes simplified explanations, examples, and references to better support the learning process \citep{wolfe2008annotations, landrum2014scaffolding}. 
These models often coexist and overlap in practice \citep{brossard2009critical, metcalfe2022science}, with practitioners employing multiple strategies to increase the readability of complex information.

In regular practice, two notable challenges persist for researchers when attempting to effectively communicate their findings to lay audiences.  
First, placing the burden of effective communication on scholars, who are trained in conducting research and not in effective communication, can be overwhelming. 
Second, translating scientific findings that heavily utilize specialized jargon often involves sacrificing some specificity in favor of accessibility. 
Balancing the precision required by the scientific community with the clarity needed for public understanding remains a significant challenge.
However, in order to uphold the fundamental principles of open science and the goal of science communication, the public's understanding of scholarship should be considered a requisite part of the scholarly process \citep{burns2003science, fecher2014open}.

\subsection{Text Simplification}
The natural language processing community addresses document accessibility through automated \emph{text simplification}, a process that rewrites complex documents using simpler grammar and vocabulary while retaining the original meaning \citep{chandrasekar1996motivations, alva2020data, al2021automated}.
In the early days of text simplification, knowledge-based approaches were widely used \citep{de2010text, zhao2018integrating, hijazi2022grass}. 
These methods relied on predefined sets of linguistic rules to transform complex sentences into simpler forms. 
For example, such systems would apply syntactic transformations, such as breaking down compound sentences into simpler ones, and lexical substitutions, where difficult words were replaced with more frequent, simpler synonyms.
While effective in specific cases, rule-based systems were limited by the rigidity of their rules, often struggling with the variability of natural language.

With the rise of machine learning, approaches that learn simplification patterns from large corpora of text have become prominent.
The most common method currently is fine-tuning a pre-trained language model with parallel corpora containing paired documents of different readability levels \citep{xu2015problems, goldsack2022making, joseph2023multilingual}.
These parallel corpora allow the model to ``translate'' complex texts into simpler ones, typically using cross-entropy loss to align the predicted simplified version with the target text. 
This shift to data-driven methods allows for greater flexibility and more nuanced simplifications, as the models learn from a broader range of linguistic patterns.

The effectiveness that supervised fine-tuning can achieve depends on the quality of the parallel corpus, specifically whether the simplified documents are sufficiently accessible.
As an example, \citet{devaraj2021paragraph} achieved approximately a one-grade reading level improvement, comparable to the readability of the plain-language samples, by fine-tuning a language model (i.e., BART \citep{lewis2019bart}) with a corpus of pairs of technical and plain-language medical texts.
This is expected: for sequence-to-sequence models, such as BART and T5, the objective of supervised fine-tuning is to minimize the Kullback–Leibler (KL) divergence between the predicted token distribution and the target distribution. 
Similarly, for causal language models, such as GPT and Gemma, minimizing KL divergence is equivalent to maximizing the joint probability of generating accessible sequences, conditioned on the abstracts.
Consequently, the best outcome that models can achieve through supervised fine-tuning is to yield lay versions that as simple as---\emph{but no simpler than}---the simplified examples in the corpus.

The problems is that, in practice, high-quality corpora for simplification are either unavailable or prohibitively expensive to obtain, and widely adopted text simplification targets are often only marginally more readable than their sources.
For example, the Simple English Wikipedia is, on average, less than two grade levels more readable than standard English Wikipedia articles \citep{isaksson2018}.
This issue also extends to abstract simplification tasks, where the simplified documents are typically less than one grade level more readable than the abstracts themselves \citep{wang2025simplifying}. 
\citet{carlson2018evaluating} proposed evaluating the readability of simplified texts using different versions of the Bible, which cover a wide spectrum of readability levels. 
This approach is specifically designed for evaluation purposes rather than for training a general-purpose simplification model.

Efforts to overcome this data bottleneck have focused on two main approaches.
First, scholars have modified the cross-entropy objective to achieve additional improvements, often by incorporating knowledge on word-level difficulty \citep{nishihara2019controllable, devaraj2021paragraph, yanamoto2022controllable}.
For instance, by adding pre-determined terms to the cross-entropy objective to reduce the model's probability assigned to a set of technical words---identified using a classifier trained on essays of different readability levels---a BART model can achieve an additional one-grade reading level improvement over one trained with the standard cross-entropy objective \citep{devaraj2021paragraph}.
However, while including extra terms in the objective function can be useful, this approach tends to be rather rigid and may compromise language quality.
Moreover, the performance gains from these improvements only improve readability by around one grade level.
Additional performance gains have also been observed when fine-tuning larger language models (ranging from 2 to 7 billion parameters) \citep{wang2025simplifying}.
Having been exposed to a significantly larger number of documents, the model may land on a parameter landscape slightly better than the one represented by the target documents.
Although using a larger language model results in about a three-grade-level boost in readability, it still falls approximately three grade levels short of making the abstracts generally accessible to our target audience---individuals without a college degree.

Our means of addressing the data bottleneck is to optimize language models by incorporating dynamically adjusted terms into the cross-entropy objective in an actor-critic manner.
This method aligns with reinforcement learning techniques that integrate multiple rewards from different types of feedback;
in long-form question-answering tasks, \citet{wu2023finegrainedhf} have demonstrated that incorporating various reward types---such as coherence, relevance, and completeness---outperforms approaches relying on a single, holistic reward.
We adapted this approach by assessing document readability through a careful balance of two heuristics proposed by \citet{riddell2021varieties}; these estimate syntactic and lexical difficulty, respectively.

\section{Scientific Abstract-Significance Statement (SASS) Corpus\label{sec: corpus_sass}}

\begin{figure*}[!ht]
  \centering
  \includegraphics[scale=0.65]{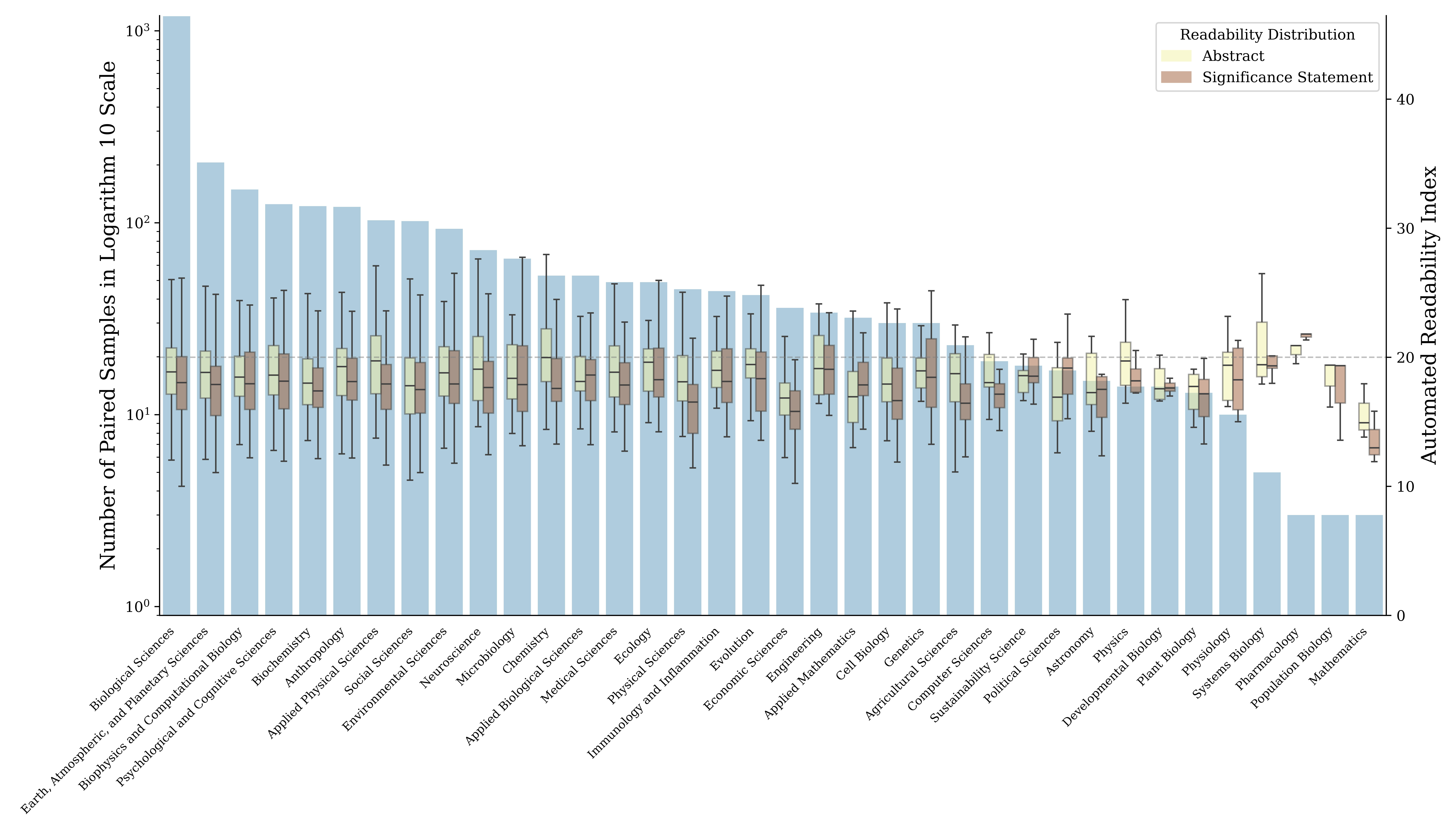}
  \caption{Discipline and readability distributions of abstracts and significance statements found in the training set of the Scientific Abstract-Significance Statement corpus. The count of paired samples in different disciplines is shown in blue bars on a log10 scale (disciplines with fewer than ten samples are not shown). Readability is measured using the Automated Readability Index (ARI), which estimates the number of years of schooling required to understand a text. On average, abstracts have a readability slightly below 20 ARI, indicating a post-graduate level. Significance statements are generally more readable than their corresponding abstracts. Orange arrows indicate the change in readability from abstracts to significance statements.\label{fig: discipline_dist}}
\end{figure*}

We used the Scientific Abstract-Significance Statement (SASS) corpus in our experiments.
This corpus is composed of 3,430 abstract-significance statement pairs derived from \emph{PNAS} and divided into training (3,030 samples), validation (200 samples), and test sets (200 samples) \citep{wang2025simplifying}. 
It covers a wide range of disciplines, ensuring diverse representation across various fields, as shown in Figure~\ref{fig: discipline_dist}. 
Corpus statistics are shown in Table~\ref{tbl: corpus_stats}; refer to Section~\ref{sec: evaluation} for a detailed description of the measures.

\begin{table*}[!ht]
\caption{
Corpus statistics for the Scientific Abstract-Significance Statement (SASS) corpus. Metrics include: ARI (Automated Readability Index), F-K (Flesch-Kincaid readability test), VOA (log ratio of proportion of words found in the VOA1500 vocabulary), SL (average sentence length and number of sentences), WA (word accessibility; log frequency per 1 billion tokens in English Wikipedia), and WL (average word length).
Measures whose names are followed by a down arrow symbol ($\downarrow$) indicate that lower values correspond to a more readable document.
Numeric values in parentheses are the corresponding standard deviations. 
Paired t-tests were conducted for each metric comparing the abstracts and significance statements, with p-values adjusted using the Bonferroni correction for multiple comparisons.
The observed differences in each of the measurements are statistically significant after adjusting for the grouped p-values at a significance level of 0.05.
\label{tbl: corpus_stats}
}
\small
\centering
\begin{tabular}{p{0.1\linewidth}p{0.075\linewidth}p{0.07\linewidth}p{0.075\linewidth}p{0.075\linewidth}p{0.075\linewidth}p{0.075\linewidth}p{0.075\linewidth}p{0.075\linewidth}}
\toprule
    Section & ARI$\downarrow$ & F-K$\downarrow$ & VOA & SL$\downarrow$ & WA & WL$\downarrow$   \\ \midrule
    Abstract & \scriptsize 18.9 \tiny (2.8) & \scriptsize 19.2 \tiny (2.4) & \scriptsize -0.43 \tiny (0.25) & \scriptsize 25.4 \tiny (4.9) & \scriptsize 12.0 \tiny (0.4) & \scriptsize 5.3 \tiny (0.4)  \\ \cmidrule{1-7}
    Significance & \scriptsize  18.1* \tiny (3.1)  & \scriptsize 18.6* \tiny (2.7) & \scriptsize -0.31* \tiny (0.26) & \scriptsize 23.9* \tiny (5.3)  & \scriptsize 11.9* \tiny (0.4) & \scriptsize 5.4* \tiny (0.4)  \\ 
\bottomrule
\end{tabular}
\end{table*}

We observed that significance statements are semantically coherent with their corresponding abstracts.
The corpus statistics indicate that significance statements are more readable than abstracts, as shown by lower mean values in the Automated Readability Index (ARI) and Flesch-Kincaid readability test (F-K). 
This suggests that the SASS corpus can be useful in simplifying scholarly abstracts across diverse disciplines.

We also observed that word accessibility (i.e., log frequency per 1 billion tokens found in English Wikipedia) and average word length suggest that significance statements can be less accessible at the word level than are their corresponding abstracts. 
Although the log ratio of words found in the VOA1500 vocabulary is slightly lower than in the corresponding abstracts, these 1,500 words are very basic and include a high proportion of function words. 
Considering that significance statements use approximately 1.5 fewer words on average, the increased use of VOA words may be a consequence of the higher use of function words to maintain grammaticality.

\section{Reinforcement Learning from Accessibility Measures}\label{sec: RLAM}
\subsection{Language Modeling via Proximal Policy Optimization}
At the core of our approach is language modeling with Proximal Policy Optimization (PPO) \citep{schulman2017proximal} guided by two accessibility measures. 
A causal language model trained on large corpora can generate the next token based on the current sequence, which is useful in the context of reinforcement learning for developing a policy model that determines the most appropriate next token to maximize the expected return in terms of document readability.

The process begins with an input sequence $s_0 = (a_0, a_1, \ldots, a_i)$, where each $a_i$ is from a set of tokens $W$, and $s_0$ represents an abstract formatted in a simple template.\footnote{The input reads: ``\texttt{Rewrite this abstract in plain English for middle school students: \{abstract\}\textbackslash n Lay summary: \{significance\}.}''}
The language model $\pi_\theta$ then generates $a_0, a_1, \ldots, a_{T-1} \sim \pi_\theta(\cdot \mid s_t)$, creating its accessible version until the maximum number of tokens $T$ is reached, either due to the context length or an end-of-sentence token:

\begin{equation}\label{eq: language_modeling_formula}
\pi_{\theta}(a_0, a_1, \ldots, a_{T-1}) = \prod_{t=0}^{T-1} \pi_{\theta}(a_t \mid s_t)
\end{equation}

Our objective is to learn a policy model that, given an abstract, models the joint probability of tokens leading to a high reward in terms of accessibility while maintaining semantic coherence. 
Formally, this is expressed as:

\begin{equation}\label{eq: objective}
J(\pi_\theta) = \mathbb{E}_{\tau \sim \pi_\theta} \left[ \sum_{t=0}^{T-1} \left( r(s_t, a_t) - \beta_{\text{KL}} \text{KL}(\pi_{\theta}(a_t \mid s_t) \parallel \pi_{\theta_{\text{SFT}}}(a_t \mid s_t)) \right) \right]
\end{equation}

Here, $J(\pi_\theta)$ represents the expected return when following policy $\pi_\theta$. The reward $r(s_t, a_t)$ is estimated for each time step $t$ in the trajectory $\tau = (s_0, a_0, s_1, a_1, \ldots, s_{T-1}, a_{T-1})$, where $s_t$ is the sequence of tokens at time $t$, formed as the concatenation of $s_0$ and the tokens $a_0, a_1, \ldots, a_{t-1}$. 
This formula iteratively computes the rewards given the current sequence $s_t$ and the token $a_t$ chosen by the policy. 
The $\beta_{\text{KL}}$-weighted KL divergence term $\text{KL}(\pi_{\theta}(a_t \mid s_t) \parallel \pi_{\theta_{\text{SFT}}}(a_t \mid s_t))$ is applied at every step of sequence generation to ensure the policy does not deviate significantly from the supervised fine-tuned model.
This is crucial because, without such a constraint, the policy model might quickly learn to output whatever the reward model favors to maximize its return, which can lead to undesirable behaviors. For instance, the model might repeatedly output a frequent word (``is is is ...''), achieving a high reward based solely on accessibility measures but lacking meaningful content.
Following \citet{stiennon2020learning}, $\beta_{\text{KL}}$ is dynamically adjusted by targeting a specific KL divergence between $\pi_{\theta}$ and $\pi_{\theta_{\text{SFT}}}$ using a capped proportional controller in logarithmic space.
The benefit of using a dynamic KL control, as opposed to a fixed one, is that it allows the model to adapt more flexibly to different stages of training, accommodating varying levels of KL divergence between the policy and SFT model.
For further details, see Appendix~\ref{appdx: ppo} for the specific PPO implementation we are using and Appendix~\ref{appdx: adapative_kl_controller} for the dynamic proportional controller.

\subsection{Reward Function}\label{sec: reward_function}
The reward function evaluates the overall quality of the output ($s_T$). 
After the initial failures of testing a traditional readability measure (i.e., ARI) as the criterion, we decided to use a balance of two accessibility measures: average sentence length in words and word accessibility, adopted from \citet{riddell2021varieties}, to guide the optimization. 

\paragraph{Word Accessibility Reward}
A word's accessibility is approximated by how frequently it appears in a large reference corpus. 
We chose the English Wikipedia corpus due to its domain similarity and applied a Moses tokenizer, yielding a vocabulary of 14.6 million types from a total of 3.6 billion tokens. 
If a token is among the most common 100,000 types, we report its frequency per billion tokens as its accessibility measure.
Otherwise, we estimate its frequency using ridge regression with an $\ell2$-norm coefficient equal to $1.0$. 
This model allows us to make serviceable estimates of the frequency of arbitrary tokens, including tokens that do not appear in the reference corpus. 
This model takes as input the token's length in Unicode code points, its byte unigrams, byte bigrams, and byte trigrams. 
The model estimates the token's log frequency per 1 billion tokens. 
We used the natural logarithm of frequencies per billion tokens as the measure of word accessibility.\footnote{We have faithfully followed the experiment of \citet{riddell2021varieties} with three differences. 
First, our reference corpus is the English Wikipedia, whereas the original study used the Common Crawl News corpus. 
Second, we did not discard duplicated sentences as \citet{riddell2021varieties} did, because we found that sentence duplication is not common in Wikipedia. 
Third, the original study reported word \emph{inaccessibility} scores by negating the logarithm of frequency per billion. We report \emph{accessibility}, without negation, because it is more naturally suited to serve as a reward. Refer to \citet[pp. 1186--1187]{riddell2021varieties} for the training of the ridge regression.} 
For example, the accessibility score for ``big'' is 11.8, while ``colossal'' scores 7.3. 
Despite being comparable in meaning, the model's production of the latter will receive fewer rewards.
Coefficient $\beta_\text{{wa}}$ is to control the scale of the credit given for word accessibility.

\paragraph{Sentence Length Reward}
Sentence length is also determined by a Moses tokenizer, which preserves hyphenation and splits contractions.\footnote{For the Moses rule-based tokenizer, we use the \texttt{sacremoses} Python package.} 
We negate the value of sentence length for intuitive calculation of the rewards for optimization.

\section{Experiment Setup}
\subsection{Training}\label{sec: training}
We initialized the policy models $\pi_{\theta}$ by adopting the Gemma-2B checkpoint reported by \citet{wang2025simplifying} ($\pi_{\theta_{\text{SFT}}}$). 
The original Gemma-2B was trained on three trillion tokens, consisting of publicly available data as well as proprietary datasets comprising ``primarily English data from web documents'' \citep{gemmateam2024gemma}.
The specific checkpoint we adopted was fine-tuned using the SASS corpus in a straightforward manner. 
It was chosen for its strong performance in simplification quality, its faithfulness, and its relatively compact size.

The two accessibility rewards were weighted as follows: the word accessibility reward was set to $\beta_{\text{WA}} = 4.0$, and we report four models with gradually increasing $\beta_{\text{SL}}$ values of $0.05$, $0.08$, $0.1$, and $0.2$.\footnote{Selecting hyperparameters for PPO training is a computationally intensive task, particularly due to the sensitivity of reward model weights. 
We grounded our search in intuitive defaults commonly adopted in PPO literature---learning rate, batch size, and value function coefficient---and verified training stability on a surrogate task (IMDb sentiment alteration). 
After confirming reliable PPO dynamics, we conducted a targeted random sweep over $\beta_{WA}$ and found that values near $\beta_{WA} = 4.0$ consistently produced stable learning and favorable trade-offs between readability and semantic retention. 
We then refined the joint configuration of $\beta_{KL}$ and $\beta_{WA}$ to further stabilize optimization and prevent collapse and semantic drift.
While we acknowledge that this hyperparameter configuration may not be globally optimal, it reflects a deliberate balance between simplicity and effectiveness. 
Rather than relying on a separate language model trained with extensive human annotations to approximate human evaluation \citep{ouyang2022training}, our reward function---implemented in just a few lines of Python---offers an elegant and computationally efficient solution for guiding the simplification process.
In addition, because word accessibility is in logarithmic space, we subtracted 10 from the estimated word accessibility and reset any values lower than 10 to 0 to keep them within a reasonable range.}
For adaptive control of the per-token semantic reward, we started with an initial $\beta_{\text{KL}} = 0.2$ and targeted a KL divergence of $8.0$ nats during the training course, capping it in the range between $0.15$ and $0.25$. 
We used a micro batch size of $4$, with each sequence used to run the PPO algorithm for $4$ epochs using importance sampling, with gradient accumulation steps set to $4$. 
We used a clip range of $0.2$ for the policy gradient and value function estimation to ensure stability.
The value function coefficient was set to $0.1$.
The optimization used standard AdamW optimizer parameters ($\beta_1 = 0.9$, $\beta_2 = 0.999$, $\epsilon = 1 \times 10^{-8}$) \citep{kingma2014adam, loshchilov2018decoupled}.
The learning rate was fixed at $1 \times 10^{-6}$.
The training was conducted on 2 H100 (80GB) GPUs using mixed precision training in bfloat16. 
The sampling temperature was set to $0.7$ in the rollout phase.
Following \citet{huang2024n}, we assigned unfinished roll-outs (indicated by the missing end of sequence token) with a fixed low score to encourage the model to generate complete simplified narratives.
We also report a model trained as dictated by ARI\footnote{We began with pilot experiments using ARI as the sole reward, and observed that it encouraged the model to take shortcuts, often collapsing outputs into overly compressed summaries---sometimes as short as a single sentence. 
While such ultra-brief abstracts resemble the desired behavior in extreme summarization tasks \citep{narayan2018dont}, they proved suboptimal for enhancing accessibility. 
These results underscored the need to explicitly guard against excessive sentence shortening; without such constraints, reinforcement learning risks compromising the semantic fidelity of the output.} (i.e., RLARI; described in Section~\ref{sec: simplification_challenges}), with all other parameters kept the same as those guided by the accessibility measures.

We performed multiple runs for each reinforcement learning process and selected the checkpoint to report based on a balance of semantic retention and ARI score obtained on the validation set. 
We observed that the readability of the generated text on the validation set often began to decline rapidly after plateauing for a while, typically accompanied by a surge in the standard deviation of average word accessibility among sentences.
This signals that the language model was sacrificing language quality and semantic relevance for extra improvements in readability. Therefore, we reported the checkpoint immediately before such instability occurred.

\subsection{Evaluation}\label{sec: evaluation}
We evaluated the simplified texts generated by the language models trained with the reinforcement learning framework using 200 abstracts from the SASS corpus test set for simplification.
Though advanced decoding methods might further refine the quality of the outputs, we used multinomial sampling with the temperature set to zero to intentionally produce the most deterministic outputs. 
This approach helped us better understand the modeling of accessible language and made it easier for us to identify potential quirks.
We assessed the quality of the generated simplified abstracts both quantitatively and qualitatively.
Quantitatively, we measured the generated texts based on their semantic retention and accessibility using established relevance measures as well as readability and accessibility metrics.

\subsubsection{Semantic Retention}
BERTScore calculates the cosine similarity between each token in the candidate sentence and each token in the reference sentence using contextual embeddings from a pre-trained language model; its results align well with human judgment on semantic similarity evaluation \citep{zhang2020bertscore}. 
It is not directly influenced by lexical overlap, making it more suitable for evaluating simplification systems than are metrics that rely on matching words, such as BLEU \citep{papineni2002bleu}.
For our evaluation, we used embeddings from the 18th layer of a BERT-large-uncased model and reported the F1 score. 
This choice is based on prior findings indicating that the 18th layer yields a strong Pearson correlation (0.72) on the WMT16 To-English benchmark \citep{zhang2020bertscore}.

\subsubsection{Simplification \& Accessibility}\label{eval: accessibility}
Accessibility can be measured with respect to the overall simplification quality (SARI); readability (ARI and Flesch-Kincaid); and other straightforward document complexity measures, including average sentence length, word accessibility (i.e., log frequency per 1 billion tokens found in English Wikipedia), the log ratio of its proportion of VOA Special English words (1,517 types in total), and average word length.

SARI (System output Against References and against the Input sentence) is specifically designed to evaluate text simplification \citep{xu2016sari}. 
It aims to measure how well a simplified text retains the original meaning while improving readability. 
SARI provides a balanced measure of how well a text simplification system performs by focusing on the necessary operations of adding, deleting, and retaining words. 

ARI and Flesch-Kincaid readability tests assign a numerical score to text that reflects the U.S. grade level required for comprehension.
Lower scores (1--13) indicate content suitable for kindergarten through twelfth grade, with each score corresponding to a subsequent grade level. Scores in the range of 14--18 suggest college-level readability, ranging from first- to senior-year content appropriateness. 
Higher scores (19 and above) are associated with advanced college education. 
Both measures use average sentence length.
Flesch-Kincaid uses syllables per word, while ARI uses characters per word for its linear combination with sentence length.

We harvested VOA Special English vocabulary comprising 1,517 unique words (VOA1500).\footnote{We included VOA Special English Word Book Sections A-Z, Science Programs, and Organs of the Body hosted on Wikipedia (\url{https://simple.wikipedia.org/wiki/Wikipedia:VOA_Special_English_Word_Book}).}
We calculate the ratio of words that appear in the VOA1500 to those that do not, then report the natural logarithm of this ratio for each generated sample.
Values above 0 indicate that the text contains more Special English words than non-Special English words, and a higher value indicates a greater presence of ``easy'' words.
We chose the VOA word list because it is a well-known, publicly available vocabulary list specifically designed for learners of English as a second language. 
Its use aligns with established heuristics in readability research, which suggest that texts using learner-friendly words tend to be more accessible.

\subsubsection{Language Quality, Faithfulness, \& Completeness}
We manually examined 5\% of all generated samples, corresponding to a randomly chosen subset of the test set from the SASS corpus.
Each generation is annotated with respect to language quality, faithfulness, and completeness using a rubric of Good, Acceptable, and Poor.
We focused on fluency and grammaticality and hand-picked both good and problematic examples when evaluating language quality.
For the evaluation of faithfulness, we conduct close readings to assess the extent to which a simplified abstract remains factually faithful to the original narrative. 
If uncertainty arises, we consult the corresponding manuscript, as our abstract simplification system must avoid producing misinformation.
Completeness is also a key consideration, as it is essential to include the main findings and implications of the research for the general public, since this is the primary goal of scientific dissemination.

\subsubsection{Token Distribution Shift Analysis}
To understand the impact of optimization guided by different rewards, we analyzed the shifts in token distribution in generations from reinforcement learning models compared to those generated by the supervised fine-tuning model from which they were initiated. 
Following the approach of \citet{lin2023unlocking}, we first generated a response using a model trained with reinforcement learning for a given query. 
We then used the supervised fine-tuning model to predict the most probable token at each position based on the context up to that point.

Token positions were categorized into three groups based on how their ranks shifted between the supervised fine-tuning model and the reinforcement learning-optimized models. 
The first group, unshifted positions, includes tokens that maintained their top rank in both models. 
The second group, marginal positions, consists of tokens that dropped slightly in rank, appearing as the second or third choice in the supervised fine-tuning model. 
Finally, shifted positions are those where a token's rank fell outside the top three choices in the supervised fine-tuning model. 
This categorization allows us to identify which tokens are affected, thereby understanding the impact of reinforcement learning guided by different rewards on the model's decision-making process.

\section{Findings \& Discussion}\label{sec: findings}

\subsection{Quantitative Assessment}
Table~\ref{tbl: automatic_evaluation} summarizes the performance of Gemma-2B, tuned in different ways, when evaluated on the test set of the SASS corpus.
The first scenario is the supervised fine-tuned baseline (SFT), which performs next-token prediction on the SASS corpus training set.
The second and third scenarios are reinforcement learning through PPO guided by ARI (RLARI) or accessibility measures (RLAM).
We assessed the generation quality by considering both semantic retention and simplification, specifically using BERT score (BS), SARI, ARI, Flesch-Kincaid readability test (F-K), the log ratio of words in the VOA1500 vocabulary (VOA), sentence length (SL), word accessibility (WA), and word length (WL).
A one-tailed paired t-test was conducted for each metric to compare observations between the reinforcement learning and supervised fine-tuning baselines, assuming improvement in document readability.
Bonferroni correction was applied to each set of tests to maintain a family-wise significance level of 0.05.

\begin{table*}[!ht]
\caption{
Comparison of Gemma-2B's performance across three approaches: the supervised fine-tuned baseline (SFT), reinforcement learning guided by ARI (using an intermediate checkpoint before significant policy gradient instability was observed, RLARI), and reinforcement learning guided by two accessibility measures (RLAM). SFT was fine-tuned using the Scientific Abstract-Significance Statement (SASS) corpus reported in \citet{wang2025simplifying}. The columns labeled $\beta_{\text{WA}}$ and $\beta_{\text{SL}}$ pertain specifically to RLAM, where the rewards for average word accessibility and sentence length are balanced.
The inference on the test split from SASS uses multinomial sampling. Metrics ARI, F-K, SARI, VOA, SL, WA, WL, and BS stand for Automated Readability Index, Flesch-Kincaid readability test, log ratio of VOA1500 vocabulary, sentence length, word accessibility, word length, and BERTScore (F1), respectively. Measures followed by a down arrow symbol ($\downarrow$) indicate that lower values are better. Numeric values in parentheses are the corresponding standard deviations. A paired two-tailed t-test was performed on observations of each measure between each model and the original abstracts. 
At a model-wise p-value of 0.05, measures that differ significantly from the SFT baseline are marked with an asterisk.
\label{tbl: automatic_evaluation}}
 \small
 \centering
 \begin{tabular}{p{0.08\linewidth}p{0.03\linewidth}p{0.03\linewidth}p{0.07\linewidth}p{0.07\linewidth}p{0.07\linewidth}p{0.08\linewidth}p{0.07\linewidth}p{0.07\linewidth}p{0.07\linewidth}p{0.07\linewidth}}
    \toprule
        Model & $\beta_{\text{SL}}$ & $\beta_{\text{WA}}$  & ARI$\downarrow$ & F-K$\downarrow$ & SARI & VOA & SL$\downarrow$ & WA & WL$\downarrow$ & BS  \\ \midrule
        SFT & \scriptsize - & \scriptsize -   & \scriptsize 15.5 \tiny (3.0) & \scriptsize 16.5 \tiny (2.6) & \scriptsize 39.1 \tiny (5.0) & \scriptsize -0.26 \tiny (0.30) & \scriptsize 20.6 \tiny (4.1)  & \scriptsize 11.9 \tiny (0.5) & \scriptsize 5.2 \tiny (0.4) & \scriptsize 0.64 \tiny (0.06)\\  
        \cmidrule{1-11}
        RLARI & \scriptsize - & \scriptsize -   & \scriptsize 12.6* \tiny (2.9) & \scriptsize 14.3* \tiny (2.5) & \scriptsize 40.1* \tiny (4.8) & \scriptsize -0.17* \tiny (0.31) & \scriptsize 16.4* \tiny (3.7)  & \scriptsize 12.0 \tiny (0.5) & \scriptsize 5.0* \tiny (0.4) & \scriptsize 0.64 \tiny (0.05)\\  
        \cmidrule{1-11}
        RLAM & \scriptsize 0.05 &  \scriptsize 4.0  & \scriptsize 13.5* \tiny (2.8) & \scriptsize 14.8* \tiny (2.4) & \scriptsize 39.8 \tiny (5.1) & \scriptsize 0.08* \tiny (0.29) & \scriptsize 21.0 \tiny (4.2)  & \scriptsize 12.7* \tiny (0.5) & \scriptsize 4.8* \tiny (0.4) & \scriptsize 0.62 \tiny (0.06)\\  
        \cmidrule{1-11}
        RLAM & \scriptsize 0.08 & \scriptsize 4.0  & \scriptsize 13.7* \tiny (3.0) & \scriptsize 14.9* \tiny (2.6) & \scriptsize 40.4* \tiny (5.2) & \scriptsize -0.02* \tiny (0.28) & \scriptsize 20.2 \tiny (3.7)  & \scriptsize 12.5* \tiny (0.5) & \scriptsize 4.9* \tiny (0.4) & \scriptsize 0.64 \tiny (0.05)\\  
        \cmidrule{1-11}
        RLAM & \scriptsize 0.1 & \scriptsize 4.0  & \scriptsize 13.3* \tiny (2.6) & \scriptsize 14.7* \tiny (2.3) & \scriptsize 40.3 \tiny (5.0) & \scriptsize -0.02* \tiny (0.31) & \scriptsize 19.3* \tiny (3.6)  & \scriptsize 12.4* \tiny (0.5) & \scriptsize 4.9* \tiny (0.4) & \scriptsize 0.64 \tiny (0.05)\\  
        \cmidrule{1-11}
        RLAM & \scriptsize 0.2 & \scriptsize 4.0  & \scriptsize 12.5* \tiny (2.9) & \scriptsize 14.0* \tiny (2.5) & \scriptsize 39.8 \tiny (5.0) & \scriptsize -0.01* \tiny (0.32) & \scriptsize 17.7* \tiny (3.4)  & \scriptsize 12.4* \tiny (0.5) & \scriptsize 4.9* \tiny (0.4) & \scriptsize 0.63 \tiny (0.05)\\  
    \bottomrule
  \end{tabular}
\end{table*}

We observe that reinforcement learning models trained with different rewards exhibit a notable reduction in reading level, bringing abstracts down to high school levels. 
The model directly guided by ARI (RLARI) achieves an ARI of 12.6, while the most performant model guided by accessibility measures (RLAM, $\beta_{\text{SL}}=4.0$ and $\beta_{\text{WA}}=0.2$) reaches 12.5, both aligning with the readability level expected for individuals who have completed K-12 education (approximately ARI 13).
However, RLARI and RLAM models achieve these readability improvements in different ways. 
For RLAM models, better readability is achieved mostly through improved token-level accessibility.
RLAM models show an increase in word accessibility from 0.5 to 0.8 compared to the supervised baseline.
This increase in the natural logarithm suggests that words generated by RLAM models are, on average, 1.6 to 2.2 times more frequent in the English Wikipedia corpus than those generated by the SFT model.
In comparison, RLARI's 0.1-unit increase in word accessibility, although observed, does not result in a statistically significant change in word frequency compared to the SFT model.
Similarly, the log ratio of the VOA1500 vocabulary in the RLAM models shows a significant improvement, with log ratios ranging from $-$0.02 to 0.08.
This implies that for every 100 non-VOA1500 (or more complex) words generated, RLAM models can produce approximately 98 to 106 VOA1500 basic words.
In contrast, the SFT and RLARI models exhibit VOA log ratios of $-$0.26 and $-$0.17, respectively, indicating that for every 100 non-VOA1500 words generated, these models produce only around 77 (84) VOA1500 words.
The average word length in characters for RLAM models ranges from 4.8 to 4.9, slightly shorter than RLARI's and outperforming SFT.
Overall, the above evidence suggests that RLAM models achieve better readability by using more common, simpler, and shorter words.

On the other hand, the RLARI model achieves better readability by producing much shorter sentences, with only a marginal boost in word-level accessibility. 
The RLARI model has the shortest average sentence length of 16.4 words, significantly outperforming the SFT model.
In comparison, a significantly shorter average sentence length is only observed when RLAM's sentence length reward coefficient ($\beta_{\text{SL}}$) exceeds 0.08.
We also observed that, in pilot studies, if $\beta_{\text{SL}}$ is set to a higher value, such as 0.5, the model's optimization will collapse after only a few hundred steps, similar to what we consistently observed in the optimization of RLARI models.
The improvement in the RLARI model's word-level accessibility is inconsistent: while we observed significant gains in the VOA log ratio and word length, the word accessibility measure did not show statistically significant improvement after a Bonferroni correction.
That said, although RLARI uses more VOA basic words and shorter words, their frequency in the English Wikipedia corpus is not high enough to result in a significant increase in word accessibility compared to the SFT baseline.

For semantic coherence with the human-written significance statements, the BERT scores from the reinforcement learning models are not significantly worse than those of the SFT model. 
This finding suggests that the generations from the reinforcement learning models are likely to remain semantically faithful and that the reinforcement learning process does not significantly degrade language quality, at least up to the point before unstable optimization signals were observed.
SARI scores from the reinforcement learning models are significantly, yet only marginally, better than those of the SFT baseline. 
Since SARI is a measurement of a combination of deletions, additions, and retention operations, a straightforward explanation is not available. 
However, the higher SARI scores confirm the greater simplification quality of the reinforcement learning models, which is likely due to a combination of simpler words and shorter sentences.

\subsection{Qualitative Analysis}

\begin{figure*}[!ht]
  \centering
  \includegraphics[scale=0.35]{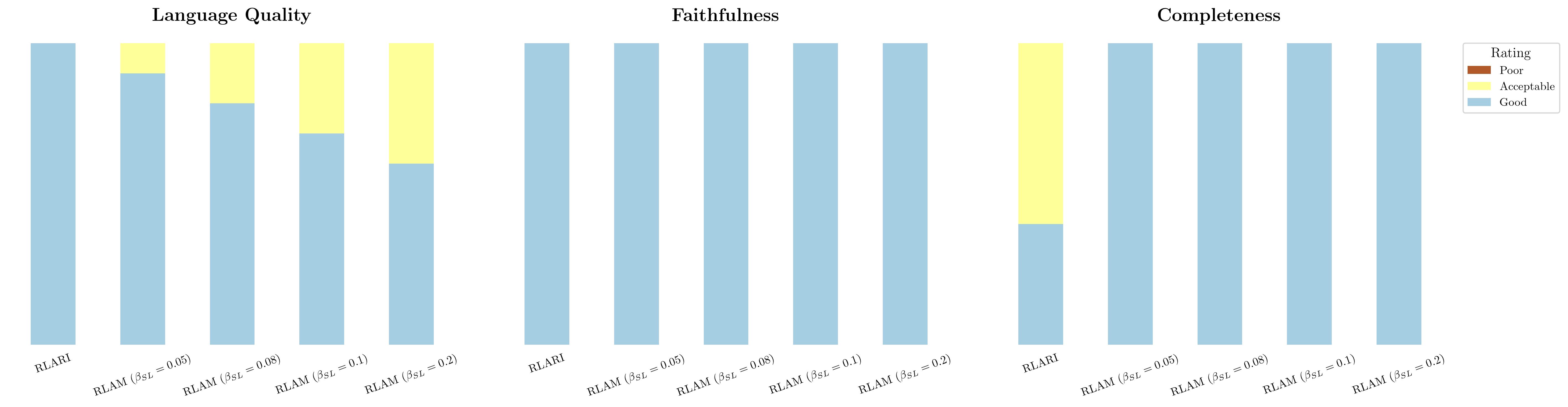}
  \caption{Results from the annotation of 5\% of the generated outputs from reinforcement learning models guided by ARI (RLARI) and accessibility measures (RLAM, with $\beta_{\text{WA}} = 4.0$ and varying $\beta_{\text{SL}}$ values), assessing language quality, faithfulness, and completeness.}
  \label{fig: annotation}
\end{figure*}

We annotated 5\% of the generated simplified abstracts from reinforcement learning models guided by ARI (RLARI) and accessibility measures (RLAM, with $\beta_{\text{WA}} = 4.0$ and varying $\beta_{\text{SL}}$ values) with respect to language quality, faithfulness, and completeness, as shown in Figure~\ref{fig: annotation}.
The test abstracts included three from biological sciences and one each from chemistry; mathematics; evolutionary biology; environmental sciences; ecology; economic sciences; and earth, atmospheric, and planetary sciences.

Regarding overall quality, we found that reinforcement learning-trained models generally produced high-quality language. 
Compared to the SFT generations, RLAM outputs are often shorter and more semantically complete, due to the imposed token budget for newly generated content (241 tokens, the length of the longest significance statement in the training set).
The main shortcoming is the presence of small trailing phrases that often deflate readability scores, such as ``(PsycINFO Database Record),'' ``(show more),'' and ``All rights reserved.'' 
The first artifact is also found in the SFT model generations and is hypothesized to be carried over from the corpora that Gemma-2B was previously exposed to, as this pattern does not appear in the SASS corpus.
An informal review of the remaining generations suggests that this phenomenon is amplified, appearing even in samples without such trailers in the SFT generations. 
The latter two artifacts were newly observed and are caused by the reinforcement learning processes. 
Where any of these issues appear, we annotate them as ``Acceptable,'' even though the generations are otherwise fluent and grammatical.

We also found that the proportion of these artifacts steadily increases as training continues, and they are usually found in only a subset of samples across different experiments.
Further, when $\beta_{\text{SL}}$ is set to a larger value, such as 0.5, the generations tend to cheat by generating readability-deflating trailers within the first few hundred iterations of optimization.
To address this problem, we experimented with an additional heuristic---the standard deviation of sentence-level word accessibility. 
When this measure exceeds approximately 0.6, it indicates that readability-deflating trailers are beginning to accumulate. 
We had limited success using this extra reward and hypothesized that it might be too indirect for the language model to effectively learn from amid other signals that may covary with this measure.
During manual examinations across experiments, no repetition was observed, which we had previously identified as a major concern in the SFT models \citep{wang2025simplifying}.

In assessments of faithfulness, we did not find any models hallucinating in the generations we examined.
However, in informal examinations, we did find that reinforcement learning checkpoints may hallucinate by generating simple but overly hedging or short expressions when over-optimized.
Generations from RLARI-trained models often remain unfinished, but they retain the main gist of the abstract.
Although we do not observe trailer phrases in the RLARI-generated texts like those found in the RLAM models, this issue frequently arises in other RLARI runs.
Subsequent checkpoints often exhibit similar problems and tend to deteriorate rapidly once the model starts cutting corners. 
This typically occurs shortly before the training process completely fails. 
However, the reported checkpoint happens to miss this characteristic.

\subsection{Token Distribution Shift Analysis for Reinforcement Learning Models}

We annotate each token in the generations according to the three categories defined in our token distribution shift analysis. 
These categories are: unshifted positions (where the supervised fine-tuning model's top prediction agrees with that of the reinforcement learning model), marginal positions (where the reinforcement learning model's top choice is the second or third choice of the supervised fine-tuning model), and shifted positions (where the reinforcement learning model's top prediction falls outside the top three choices of the supervised fine-tuning model).
This categorization allows us to closely examine how reinforcement learning with different rewards influences the behavior of the supervised fine-tuning model from which they were initiated.

\begin{table*}[!ht]
\caption{Token distribution analysis across different RLAM Models and the RLARI Model. The table shows the counts and proportions of tokens classified as marginal, shifted, and unshifted for each model. RLAM models are evaluated with different $\beta_{\text{SL}}$ values while keeping $\beta_{\text{WA}}$ constant at 4.0. Values in parentheses represent their proportion relative to all generated tokens of the respective model.\label{tbl: tds_stats}}
 \small
 \centering
 \begin{tabular}{>{\centering\arraybackslash}p{0.08\linewidth} >{\centering\arraybackslash}p{0.03\linewidth} >{\centering\arraybackslash}p{0.03\linewidth} >{\centering\arraybackslash}p{0.14\linewidth} >{\centering\arraybackslash}p{0.12\linewidth} >{\centering\arraybackslash}p{0.14\linewidth}}
    \toprule
        Model & $\beta_{\text{SL}}$ & $\beta_{\text{WA}}$  & Marginal Tokens & Shifted Tokens  & Unshifted Tokens  \\ \midrule
        RLARI & \scriptsize - & \scriptsize -   & \scriptsize 2,303 (5.96\%) & \scriptsize 183 (0.47\%) & \scriptsize 36,128 (93.56\%) \\
        \cmidrule{1-6}
        RLAM & \scriptsize 0.05 &  \scriptsize 4.0  & \scriptsize 2,564 (10.12\%) & \scriptsize 354 (1.40\%) & \scriptsize 22,415 (88.48\%) \\  
        \cmidrule{1-6}
        RLAM & \scriptsize 0.08 & \scriptsize 4.0  & \scriptsize 2,297 (8.36\%) & \scriptsize 274 (1.00\%) & \scriptsize 24,893 (90.64\%) \\  
        \cmidrule{1-6}
        RLAM & \scriptsize 0.1 & \scriptsize 4.0  & \scriptsize 2,468 (8.00\%) & \scriptsize 311 (1.01\%) & \scriptsize 28,054 (90.90\%) \\  
        \cmidrule{1-6}
        RLAM & \scriptsize 0.2 & \scriptsize 4.0  & \scriptsize 2,406 (8.93\%) & \scriptsize 346 (1.28\%) & \scriptsize 24,176 (89.78\%) \\  
        \bottomrule
  \end{tabular}
\end{table*}

As shown in Table~\ref{tbl: tds_stats}, we quantify the distribution of token types across models and their respective proportions relative to the total token count for each model. 
The table suggests that the RLARI model modifies the SFT baseline less frequently: 0.5\% of tokens are shifted, ca. 6.0\% are marginally shifted, while approximately 94.0\% remain unaltered. 
In contrast, the RLAM models exhibit a higher proportion of both shifted (ranging from 1.0\% to 1.4\%) and marginal (8.0\% to 10.1\%) tokens: approximately twice as many shifted tokens as the RLARI model, and a noticeably increased number of marginal tokens.
We hypothesize that the differences primarily originate from how the sentence length reward is incorporated into the optimization process.
Compared to the coefficient of sentence length in the ARI formula, we assign much smaller weights to sentence length in the rewards guiding RLAM models.
Since the supervised model has already demonstrated its ability to shorten sentences, it is more reasonable to encourage the model to focus on finding accessible word alternatives for a higher reward, rather than excessively incentivizing shorter sentences.
Over-prioritizing sentence length can also exploit the reward system by nudging the model to generate benign but meaningless trailing phrases, further derailing it from producing more accessible words and causing the optimization to collapse prematurely.
Another reason for the difference in token type distribution could be the different heuristics for word choice: RLAM models are incentivized to favor more common words, while RLARI prioritizes shorter words.

\begin{figure*}[!ht]
  \centering
  \includegraphics[scale=0.6]{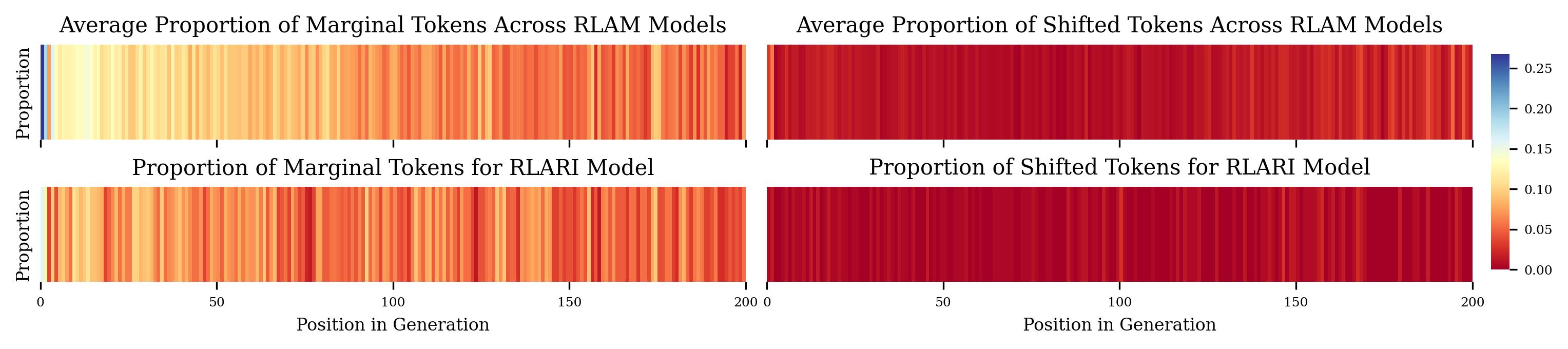}
  \caption{Token distribution shift analysis for reinforcement learning models. The figure illustrates the average distribution of marginal and shifted tokens across RLAM models (top row) and RLARI models (bottom row). The left column represents the proportion of marginal tokens, while the right column shows the proportion of shifted tokens, both relative to the total token count at each position in the generated sequences. 
  }
  \label{fig: tds}
\end{figure*}

We also analyzed the distribution of token types with respect to their positions in the generated sequences. 
In Figure~\ref{fig: tds}, the top two subplots show the average distributions from the four RLAM models for clarity. 
The left column focuses on marginal tokens, while the right column visualizes shifted tokens. 
Each subplot illustrates the proportion of each token category relative to the total token count at each position.
Our analysis reveals that the RLAM models exhibit more marginal modifications at the initial positions, whereas shifted tokens tend to accumulate toward the end of the sequences. 
This pattern suggests that, compared to the supervised fine-tuning model, which tends to overemphasize implications \citep{wang2025simplifying}, the RLAM models may truncate redundant content by employing more shifted tokens toward the end. 
Marginal shifts are used at the beginning to introduce main subjects or background information with simpler words.

In contrast, the RLARI model does not display such a systematic pattern; different types of tokens appear to be more evenly distributed across positions but lack the level of consistency observed in the RLAM models. 
We hypothesize that this difference may be caused by disproportionately large incentives for shorter sentences, which could hinder effective exploration of alternative, simpler word substitutions and ultimately contribute to the RLARI model's performance variability after the initial phases of optimization.
Additionally, approximately 90 percent of the tokens generated by the reinforcement learning models remain unchanged, which roughly corroborates previous findings in RLHF studies, where around 8 percent are changed \citep{lin2023unlocking}.
The difference may arise from the nature of the task; we focus on rephrasing in an accessible manner, which often does not require altering many tokens, unlike \citet{lin2023unlocking}, who align language models to human preferences, where the expected answer can be quite different from what would otherwise be derived from the base model.

\section{Conclusion}
To improve the accessibility of scientific literature to the general public, we implemented reinforcement learning techniques to guide language models, extending beyond the traditional cross-entropy objective.
Our study demonstrates that carefully balancing accessibility measures at the word and sentence levels can effectively guide Gemma-2B in simplifying scholarly abstracts, outperforming the supervised fine-tuning baseline by a large margin. 
This approach achieves these improvements without compromising language quality or faithfulness and mitigates the supervised fine-tuning model's tendency to overemphasize research implications.
The best model trained using our method successfully adjusts the readability level of scholarly abstracts by approximately six U.S. grade levels---in other words, from a postgraduate to a high school level.
Compared to the supervised fine-tuning model, the words generated by the model trained via our approach are proven to be more common (1.6 to 2.2 times more frequent), easier (with more VOA basic words), and shorter (by 0.3--0.4 characters).
This improvement addresses a key limitation of existing corpora, in which the target distribution (i.e., significance statements) often does not adequately prioritize the accessibility of word choice.
We also investigated the token distribution shift to better understand how reinforcement learning with different rewards influences the behavior of the SFT model from which it was initiated. 
Our analysis revealed systematic modifications in model outputs when using carefully balanced word- and sentence-level rewards, compared to traditional readability measures, which inevitably involve more aggressive rewards for reducing sentence length.

We considered the limitations of using average sentence length as a reward, as it can be a rather coarse measurement of syntactic complexity and perhaps also of cognitive comprehensibility. 
We observed that it is not uncommon for restrictive attributive clauses and other syntactic structures to appear in the generations from RLAM-trained models that could theoretically be split into separate sentences. 
However, increasing the coefficient for the sentence length reward exacerbated optimization instability, as was observed in the RLARI models.
In future work, more direct yet easily obtainable measures, such as dependency length \citep{futrell2015large}, could be a promising direction to explore for generating shorter sentences.

Another potential direction is to generate lay summaries from the full text, not just the abstract. 
With current large language models capable of handling contexts with several thousand tokens, potentially enhanced by rotary positional encoding \citep{su2024roformer}, this approach is feasible. 
This method also has the added benefit of reducing hallucinations by providing more comprehensive information from the entire manuscript. 
Also, reinforcement learning from accessibility measures is not limited to scientific literature; it can also be an effective solution for other simplification scenarios where a gold-standard corpus is unavailable. 
We hope this work contributes to bridging the gap between scholarship and a broader audience, advancing the understanding and development of better simplification systems, and ultimately fostering a more informed and engaged society.

\section*{Acknowledgments}
We gratefully acknowledge the support of the Institute of Museum and Library Services (No. RE-246450-OLS-20) and the National Social Science Fund of China (No. 23\&ZD221). 
We also thank the organizers of the LIS Education and Data Science Integrated Network Group (LEADING), including Jane Greenberg, Erik Mitchell, Kenning Arlitsch, Jonathan Wheeler, and Samantha Grabus. 
Additionally, we are thankful to Coltran Hophan-Nichols and Alexander Salois from the University Information Technology Research Cyberinfrastructure at Montana State University for providing computational resources on the Tempest High Performance Computing System, Doralyn Rossmann for research support, Michael Hartwell for proofreading, Jacob Striebel for discussions on improving sentence-level rewards, and Deanna Zarrillo for early involvement in the project.


\clearpage
\appendix

\section{Reinforcement Learning Using PPO}\label{appdx: ppo}

We fine-tune the policy model in an actor-critic manner: while the policy model (the actor) generates sequences of tokens based on the current sequence $s_t$, the critic is an additional linear layer that takes the output of the language model's last layer and produces a scalar for time step $t$, estimating the expected cumulative reward of producing the token $a_t$, noted as $V_{\rho}(a_t)$. The problem is reduced to maximizing at every step the expected cumulative reward of taking action $a_t$ in state $s_{t}$ and following the policy $\pi_\theta$ thereafter ($V_{\rho}(s_{t+1}) = V_{\rho}(s_t, a_t)$), compared to the expected cumulative reward of being in state $s$ ($V_{\rho}(s_t)$), termed as the advantage $d_t = V_{\rho}(s_{t+1}) - V_{\rho}(s_t)$.

The final reward of the entire generation (see Section~\ref{sec: reward_function}) is back-propagated through the sequence using Temporal Difference (TD) and Generalized Advantage Estimation (GAE) to estimate the advantage of each token:
\begin{equation}\label{eq: advantage_formula}
d_t = \sum_{t'=t}^{T} (\gamma \lambda)^{t'-t} \left( r_T + \gamma V_{\rho}(s_{t'+1}) - V_{\rho}(s_{t'}) \right)
\end{equation}
where $\gamma$ is the discount factor for future rewards, $\lambda$ controls the bias-variance trade-off, and $r_T$ is the final reward, which, in our case, is a linear combination of two accessibility measures.
$V_{\rho}$ is trained via minimizing a square-error loss (see Equation~\ref{eq: value_objective}).

We used the PPO \citep{schulman2017proximal} clipped surrogate objective with importance sampling to more efficiently use offline samples to update the online policy. 
Importance sampling corrects for the discrepancy between the behavior policy that generated the samples and the current policy by weighting the samples using the ratio of their probabilities under both policies. 
The PPO algorithm introduces a clipping mechanism to balance exploration and exploitation while preventing large, potentially harmful updates to the policy (see Equation~\ref{eq: ppo}).

The entire RLAM algorithm is illustrated in Algorithm~\ref{alg: rlam}.

\begin{algorithm}[H]
\caption{Training with reinforcement learning from uncombined accessibility measures. The policy model is updated using the PPO clipped surrogate objective (Eq.~\ref{eq: ppo}), and the value model is updated by minimizing a square-error objective (Eq.~\ref{eq: value_objective}).}
\label{alg: rlam}
\begin{algorithmic}[1]
\State \textbf{Input:} initial policy model $\pi_{\theta_{\text{SFT}}}$, randomly initiated value head $V_{\rho_{\text{init}}}$, reward function $R$ for the last token, weighted by $\beta_{\text{KL}}$ for KL divergence term; task prompts $X$; hyperparameters $\gamma, \lambda, \epsilon$
\State $\pi_{\theta} \gets \pi_{\theta_{\text{SFT}}}$, $V_{\rho} \gets V_{\rho_{\text{init}}}$
\For{step $= 1, \ldots, M$}
    \State Sample a batch $\{s_0\}^n$ from $X$
    \State Sample output sequences $\{a_0, a_1, \ldots, a_{T-1}\}^n \sim \pi_{\theta}(\cdot \mid s_0)$ for each prompt $s_0$ in the batch \hfill $\triangleright$ Eq.~\ref{eq: language_modeling_formula}
    \State Compute reward $r_T^n$ for each sampled sequence $\{a_0, a_1, \ldots, a_{T-1}\}^n$ using reward function R \hfill $\triangleright$ Sec.~\ref{sec: reward_function}
    \State Distribute the final reward $r_T^n$ to each token in the sequence through GAE \hfill $\triangleright$ Eq.~\ref{eq: advantage_formula}
    \State Compute advantages $\{d_t \mid s_t\}_{t=0}^{T-1}$, value targets $\{V_{\text{targ}}(s_t)\}_{t=0}^{T-1}$ for each sequence with $V_{\rho}$ and compute KL divergence penalty $\text{KL}_t = \text{KL}(\pi_{\theta}(a_t \mid s_t) \parallel \pi_{\theta_{\text{SFT}}}(a_t \mid s_t))$ 
    \For{PPO iteration $= 1, \ldots, \mu$}
        \State Update the policy model by maximizing the PPO clipped surrogate objective with KL penalty:
            \begin{align}
    \theta \gets \arg \max_{\theta} \frac{1}{n} \sum_{n=1}^{n} \frac{1}{T} \sum_{t=1}^{T} 
    &\min \left( \frac{\pi_{\theta_{\text{online}}}(a_t \mid s_t)}{\pi_{\theta_{\text{offline}}}(a_t \mid s_t)} \left( A_t - \beta_{\text{KL}} \text{KL}_t \right), \right. \nonumber \\
    &\qquad \left. \text{clip} \left( \frac{\pi_{\theta_{\text{online}}}(a_t \mid s_t)}{\pi_{\theta_{\text{offline}}}(a_t \mid s_t)}, 1-\epsilon, 1+\epsilon \right) \left( A_t - \beta_{\text{KL}} \text{KL}_t \right) \right) \label{eq: ppo}
    \end{align}
    \EndFor
    \State Update the value model by minimizing a square-error objective:
    \begin{equation}
    \rho \gets \arg \min_{\rho} \frac{1}{n} \sum_{n=1}^{n} \frac{1}{T} \sum_{t=1}^{T} \left( V_{\rho}(s_t) - V_{\text{targ}}(s_t) \right)^2 \label{eq: value_objective}
    \end{equation}
\EndFor
\State \textbf{Output:} $\pi_{\theta}$
\end{algorithmic}
\end{algorithm}

\section{Adaptive KL Controller}\label{appdx: adapative_kl_controller}

Following \citet[Sec.~2.2]{ziegler2019adaptiveKLcontrol}, we dynamically adjust $\beta_{\text{KL}}$ to target a specific KL divergence value, $\text{KL}_{\text{target}}$, using a log-space proportional controller. 

The update rule is:
$$
\beta_{\text{KL}_{t+1}} = \beta_{\text{KL}_{t}} \left(1 + K_{\beta} \cdot \text{clip}\left(\frac{\text{KL}(\pi_{\theta}, \pi_{\theta_{\text{SFT}}}) - \text{KL}_{\text{target}}}{\text{KL}_{\text{target}}}, -0.2, 0.2\right)\right)
$$
where $K_{\beta}$ is the proportional gain, set to $0.01$.

\section{Manual Evaluation}\label{appdx: manual_evaluation}

The evaluation involved three annotators: Jason Clark (academic librarian), Zuoyu Tian (computational linguist), and Haining Wang (information scientist).
The following rubrics were iteratively developed and refined over two months of model monitoring and interim generation reviews.

\begin{itemize}
    \item Language Quality: Evaluates the fluency and grammatical accuracy of the text.
    \begin{itemize}
        \item Good: Natural and free from grammatical errors; expressions flow properly.
        \item Acceptable: Natural and error-free, although it may contain formulaic endings (e.g., ``All rights reserved.'') that have minimal impact on readability.
        \item Poor: Contains grammatical errors.
    \end{itemize}
    
    \item Completeness: Assesses the coverage of the abstract's main components, including context, purpose, main findings, and conclusions.
    \begin{itemize}
        \item Good: Fully covers all four aspects without truncation.
        \item Acceptable: Covers at least three aspects, though the generation may be truncated.
        \item Poor: Covers fewer than three of the main aspects.
    \end{itemize}
    \item Faithfulness: Assesses the scientific fidelity of the simplified text to the original abstract.
    \begin{itemize}
        \item Good: Accurately reflects the main claims of the original abstract, with no factual errors.
        \item Acceptable: Maintains semantic alignment, though the claims remain unverifiable after 15 minutes of manual searching without AI assistance. No factual errors are present.
        \item Poor: Misrepresents findings, contains false claims, or includes factual errors.
    \end{itemize}

\end{itemize}

Each annotator independently reviewed the selected samples using the finalized rubrics before beginning formal annotation.
In the formal rating phase, discrepancies were resolved through discussion, referencing the original manuscripts and external sources as needed.
While our evaluations of language quality and completeness were generally consistent, assessing faithfulness proved challenging given the broad disciplinary diversity of the source abstracts.
For example, the original sentence states: \textit{``At high concentrations, Zn\textsuperscript{2+}, in addition, binds to a second site and inhibits the outward movement of the voltage sensor of Hv1.''} 
A simplified version reads: \textit{``At high levels, it also prevents the channel's parts from moving.''} 
While the simplified version is notably more accessible and preserves the core causal relationship, it remains unclear---without specialized domain expertise---whether the omission of ``outward'' specifies a subtle factual distortion (e.g., whether inward movement is also implicated).
In these instances, we prioritized semantic alignment for a general audience.
In retrospect, we acknowledge that incorporating inter-rater agreement metrics, such as Cohen's Kappa, for language quality and completeness would have enhanced our methodological rigor.

We have made all generations, including intermediate outputs from the RL training process, publicly available in our project repository.
We encourage readers and domain experts to evaluate the faithfulness of our outputs in light of their disciplinary expertise.

\bibliography{references}
\bibliographystyle{apalike}

\end{document}